\documentclass[10pt,twocolumn,letterpaper]{article}

\usepackage{standalone}
\usepackage{pdfpages}
\usepackage{iccv}
\usepackage{times}
\usepackage{epsfig}
\usepackage{graphicx}
\usepackage{amsmath}
\usepackage{amssymb}
\usepackage{algorithm}
\usepackage[noend]{algpseudocode}
\usepackage{mathtools}
\usepackage{tabularx}
\usepackage{balance}
\DeclareMathOperator*{\argmin}{argmin}
\DeclareMathOperator{\dis}{d}
\newcommand{\vect}[3]{\mathbf{#1}_{#2#3}}
\newcommand*\samethanks[1][\value{footnote}]{\footnotemark[#1]}

\usepackage[pagebackref=true,breaklinks=true,letterpaper=true,colorlinks,bookmarks=false]{hyperref}

\iccvfinalcopy 

\ificcvfinal\fi

\newcommand\javier[1]{\textcolor{blue}{}}
\newcommand\christoph[1]{\textcolor{blue}{}}
\begin{document}

\title{Efficient Learning on Point Clouds with Basis Point Sets}

\author{Sergey Prokudin \thanks{work was done during internship at Amazon}\\
Max Planck Institute for Intelligent Systems\\
T\"ubingen, Germany\\
{\tt\small sergey.prokudin@tuebingen.mpg.de}
\and
Christoph Lassner \thanks{the last two authors were equally involved}\\
Amazon\\
T\"ubingen, Germany\\
{\tt\small classner@amazon.com}
\and
Javier Romero \samethanks[2]\\
Amazon\\
Barcelona, Spain\\
{\tt\small javier@amazon.com}
}

\maketitle
\thispagestyle{empty}

\begin{abstract}

With an increased availability of 3D scanning technology, point clouds are moving into the focus of computer vision as a rich representation of everyday scenes.
However, they are hard to handle for machine learning algorithms due to their unordered structure. One common approach is to apply occupancy grid mapping, which dramatically increases the amount of data stored and at the same time loses details through discretization. Recently, deep learning models  were proposed to handle point clouds directly and achieve input permutation invariance. However, these architectures often use an increased number of parameters and are computationally inefficient. %
In this work we propose \textit{basis point sets} (BPS) as a highly efficient and fully general way to process point clouds with machine learning algorithms. The basis point set representation is a residual representation that can be computed efficiently and can be used with standard neural network architectures and other machine learning algorithms. %
Using the proposed representation as the input to a simple fully connected  network allows us to match the performance of PointNet on a shape classification task, while using three orders of magnitude less floating point operations. In a second experiment, we show how the proposed representation can be used for registering high resolution meshes to noisy 3D scans. Here, we present the first method for single-pass high-resolution mesh registration, avoiding time-consuming per-scan optimization and allowing real-time execution. 



\end{abstract}

\section{Introduction}

\begin{figure}[t]
\begin{center}
  \includegraphics[trim={3cm 0cm 5cm 0cm},clip,width=\linewidth]{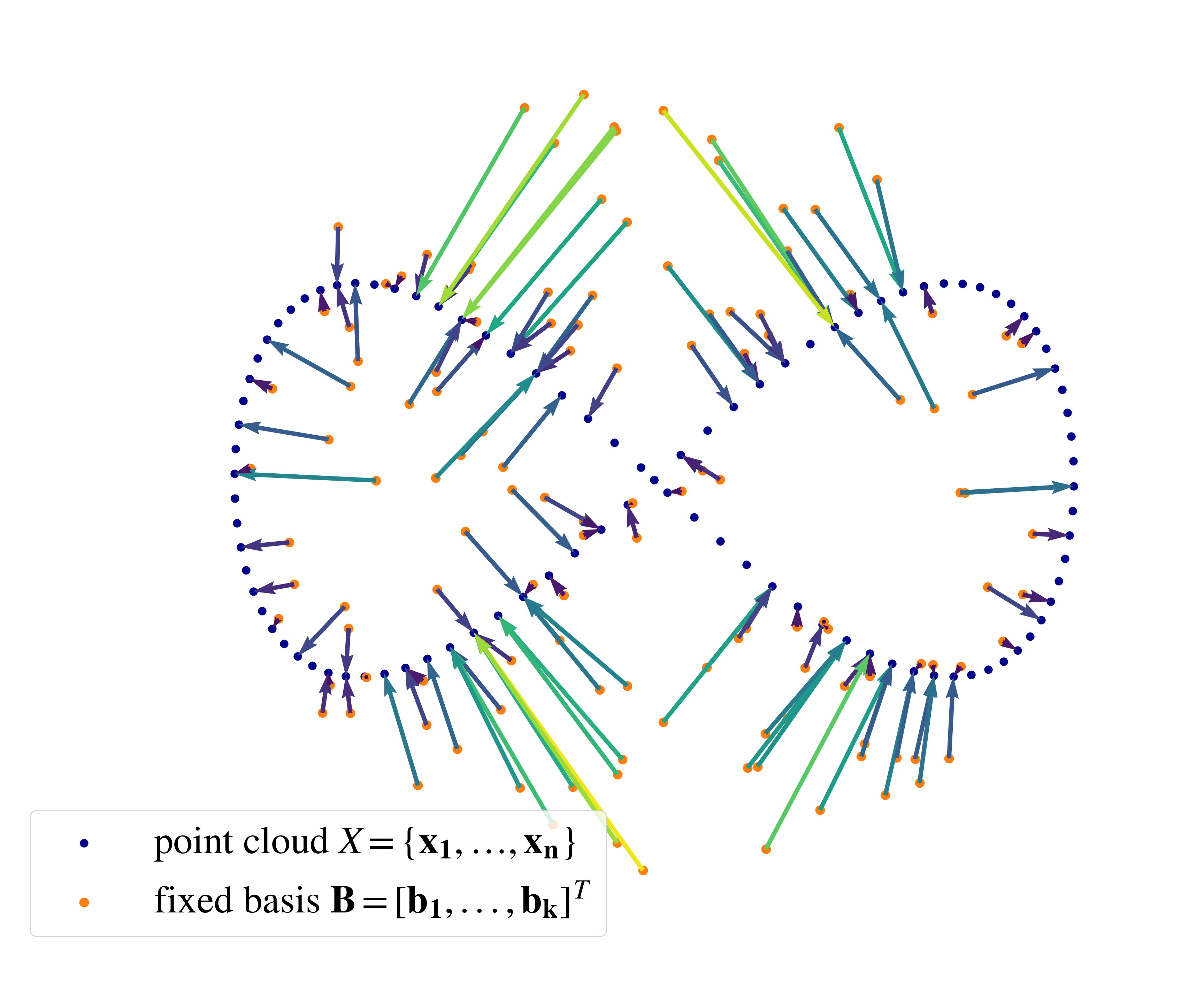}
\end{center}
  \caption{\textit{Basis point set encoding for point clouds}. The encoding of a point cloud $X=\{\mathbf{x_1}, \dots, \mathbf{x_n}\}$ is a fixed-length feature vector, computed as the minimal distances to a fixed set of points $\mathbf{B} = [\mathbf{b_1}, ..., \mathbf{b_k}]^T$. This representation can be used as input to arbitrary machine learning methods, in particular it can be used as input for off-the-shelf neural networks. This leads to substantial performance gains as compared to occupancy grid encoding or specialized neural network architectures without sacrificing the accuracy of predictions. }
\label{fig:fig1}
\end{figure}

\begin{figure*}[t!]
\centering
\includegraphics[trim={0.5cm 0cm 0.5cm 0cm},clip,width=\textwidth]{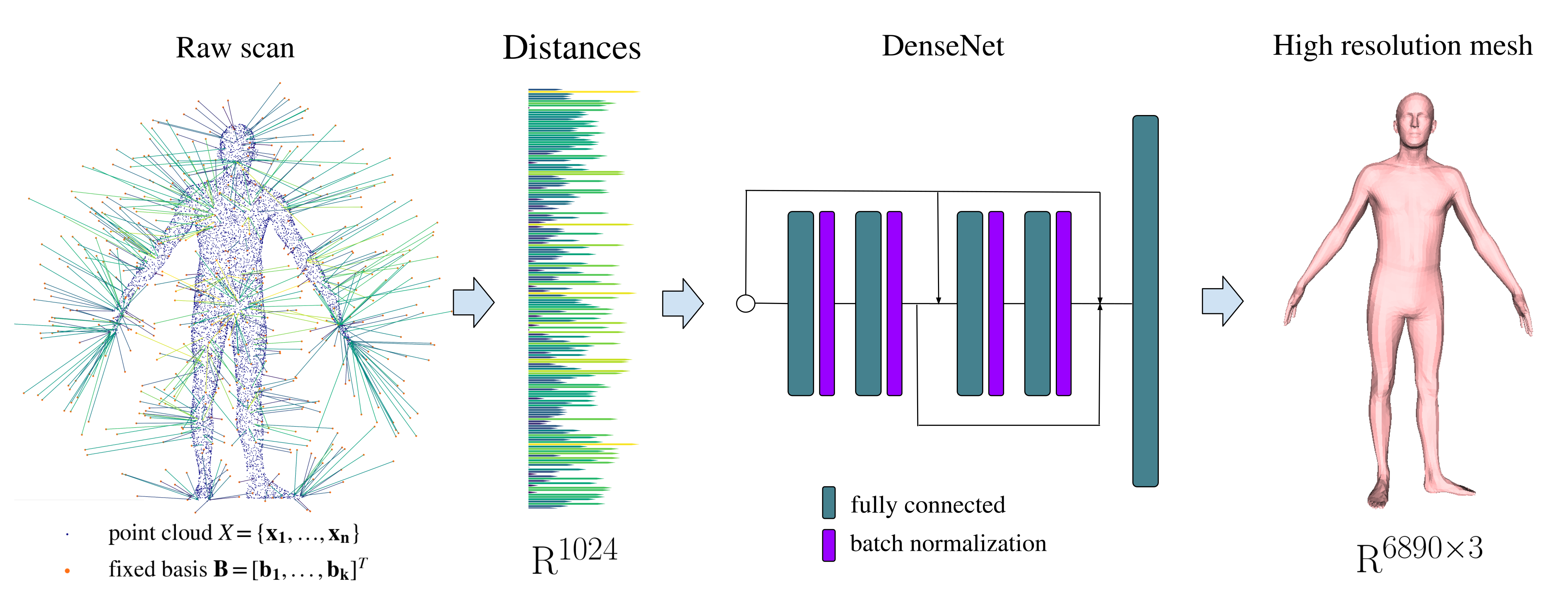}
\caption{\textit{Overview of our proposed model for the task of mesh registration to a noisy scan.} The computed minimal distances to the selected basis point set are provided as input to a simple dense network with two blocks of two fully connected layers. The model directly predicts mesh vertex positions, with a forward pass taking less than 1ms. We also propose a model for shape classification; see Sec.~\ref{section:learning} for details.}  
\label{fig:fig2}
\end{figure*}


Point cloud data is becoming more ubiquitous than ever: anyone can create a point cloud from a set of photos with easy to use photogrammetry software or capture a point cloud directly with one of many consumer-grade depth sensors available worldwide.
These sensors will soon be used in most aspects of our daily lives, with autonomous cars recording streets and city environments and VR and AR devices recording our home environment on a regular basis. The resulting data represents a great opportunity for computer vision research: it complements image data with depth information and opens up new fields of research.

However, point cloud data itself is unstructured. This leads to a variety of problems: (a) point clouds have no fixed cardinality, varying their size depending on the recorded scene. They are also not `registered' in the sense that it is not trivial to find correspondences between points across recordings of the same or of a similar scene. (b) Point clouds have no notion of neighborhood. This means that it is not clear how convolutions, one of the critical operations in deep learning, should be performed.

In this paper, we present a novel solution to the aforementioned problems, in particular the varying cloud cardinality. For an illustration, see Fig.~\ref{fig:fig1}. We propose to encode point clouds as minimal distances to a fixed set of points, which we refer to as \textit{basis point set}. This representation is vastly more efficient than classic extensive occupancy grids: it reduces every point cloud to a relatively small fixed-length vector. The vector length can be adjusted to meet computational constraints for specific applications and represents a trade-off between fidelity of the encoding and computational efficiency. Compared to other encodings of point clouds, the proposed representation also has an advantage in being more efficient with the number of values needed to preserve high frequency information of surfaces.

Given its fixed length, the presented encoding can be used with most of the standard machine learning techniques. In this paper we apply mostly artificial neural networks to build models with it, due to their popularity and accuracy.
In particular, we analyze the performance of the encoding in two applications: point cloud classification and mesh registration over noisy 3D scans (\cf, Fig.~\ref{fig:fig2}).

For point cloud classification, we achieve the same accuracy on the ModelNet40~\cite{wu20153d} shape classification  benchmark as PointNet~\cite{qi2017pointnet}, while using an order of magnitude less parameters  and three orders of magnitudes less floating point operations.
To demonstrate the versatility of the encoding, we show how it can be used for the task of mesh registration. We use the encoded vectors as input to a neural network that directly predicts mesh vertex positions. While showing competitive performance to the state-of-the-art methods on the FAUST dataset~\cite{bogo2014faust}, the main advantage of our method is the ability to produce an aligned high resolution mesh from a noisy scan in a single feed-forward pass. This can be executed in real time even on a non-GPU laptop computer, requiring no additional post-processing steps. We make our code for both presented tasks available, as well as a library for usage in other projects\footnote{\label{footnote_supp}\url{https://github.com/sergeyprokudin/bps}}. 

\christoph{Mention patent pending? Or put that on the github website?}





\section{Related Work}

\label{section:related}

In this section, we describe existing 3D data representations and models and put them in relation to the presented method. We focus on representations that are compatible with deep learning models, due to their high performance on a variety of 3D shape analysis tasks.


\textbf{Point clouds.} Numerous methods \cite{qi2017pointnet, qi2017pointnet++, shen2018mining, zaheer2017deep, li2018so} were proposed that process 3D point clouds directly, amongst which the PointNet family of models gained the most popularity. This approach processes each point separately with a small neural network followed by an aggregation step with a pooling operation to reason about the whole point cloud. Similar pooling-based approaches for achieving feature invariance on general unordered sets were proposed in other works as well \cite{zaheer2017deep}.
Other methods working directly on point clouds organize the data in kd-trees and other graphs~\cite{klokov2017escape, gadelha2017shape, landrieu2018large}. These structures define a neighborhood and thus convolution operations can be applied. Vice versa, specific convolutional filters can be designed for sparse 3d data~\cite{tatarchenko2018tangent, shen2018mining}.


We borrow several ideas from these works, such as using kNN-methods for searching efficiently through local neighborhoods or achieving order invariance through the use of pooling operations over computed distances to basis points. However, we believe that the proposed encoding and model architectures offer two main advantages over existing point cloud networks: (a) higher computational efficiency and (b) conceptually simpler, easy-to-implement algorithms that do not rely on a specific network architecture or require custom neural network layers. 


\textbf{Occupancy grids.} Similar to pixels for 2D images, occupancy grid is a natural way of encoding 3D information. Numerous deep models were proposed that work with occupancy grids as inputs~\cite{maturana2015voxnet, qi2016volumetric, moon2018v2v}. However, the main disadvantage of this encoding is their cubic complexity. This results in a high amount of data needed to accurately represent the surface. Even relatively large grids by our current memory standards ($128^3, 256^3$) are not sufficient for an accurate representation of high frequency
surfaces like human bodies.
At the same time, this type of  voxelization results in very sparse volumes when used to represent 3D surfaces: most of the volume measurements are zeros. This makes this representation an inefficient surface descriptor in multiple ways. A number of methods was proposed to overcome this problem~\cite{wang2017cnn, riegler2017octnet}. However, the problem of representing high frequency details remains, together with a large memory footprint and low computational efficiency for running convolutions.


\textbf{Signed distance fields.}  Truncated signed distance fields (TSDFs) ~\cite{curless1996volumetric, newcombe2011kinectfusion, riegler2017octnetfusion, song2017semantic, zeng20173dmatch, dai2018scan2mesh, park2019deepsdf} can be viewed as a natural extension of occupancy grids: they store distance-to-surface information in grid cells instead of a simple occupancy flag. While this partially resolves the problem of representing surface information, the cubic requirement for memory and the low computational efficiency for convolutions remains. In comparison, our method can be viewed as one that uses an arbitrary subset of points from the distance field. The crucial difference is that the distance field we sample from is unsigned and non-truncated, and the number of samples is proportional to the number of points in the original cloud. We further investigate the connection between occupancy grids, TSDFs and BPS in Sec.~\ref{subsection:comparison}.



\textbf{2D projections.} Another common strategy is to project 3D shapes to 2D surfaces and then apply standard frameworks for 2D input processing. This includes depth maps~\cite{wu20153d}, height maps~\cite{sarkar2018learning},  as well as a variety of multi-view models~\cite{su2015multi, kanezaki2018rotationnet, feng2018gvcnn}. Closely related are approaches that project 3D shapes into spheres and apply spherical convolutions to achieve rotational invariance~\cite{esteves2018learning, cohen2018spherical}. While projection-based approaches show high accuracy in discriminative tasks (classification, shape retrieval), they are fundamentally limited in representing shapes that have multiple `folds', invisible from external views. In comparison, our encoding scheme can accurately preserve surface information of objects with arbitrary topology as we show in our experiments in Sec.~\ref{section:analysis}.







We now describe the algorithm for constructing the proposed basis point representation from a given point cloud.

\section{Method}
\label{section:method}

\paragraph{Normalization.} The presented encoding algorithm takes a set of point clouds as input $\mathbf{X} = \{ X_i, i=1, \dots, p \}$. Every point cloud can have a different number of points $n_i$:

\begin{eqnarray}
X_i = \{\mathbf{x}_{i1}, \dots, \mathbf{x}_{in_i}\}, \mathbf{x}_{ij} \in \mathbb{R}^d,
\label{eq:point_cloud}
\end{eqnarray}

where $d=3$ for the case of 3D point clouds. In a first step, we normalize all point clouds to a fit a unit ball:

\begin{eqnarray}
\vect{x}{i}{j} = \frac{\vect{x}{i}{j} - \mathbb{E}_{\vect{x}{i}{j} \sim X_i}{\vect{x}{i}{j}}}{\max_{\vect{x}{i}{j} \in X_i}{\|\vect{x}{i}{j} - \mathbb{E}_{\vect{x}{i}{j} \sim X_i}{\vect{x}{i}{j}}\|}}, \forall i, j.
\label{eq:unit_norm}
\end{eqnarray}

\paragraph{\textit{BPS} construction.} Next, we form a \textit{basis point set}. For this task, we sample $k$ random points from a ball of a given radius $r$:

\begin{eqnarray}
\mathbf{B} = [\vect{b}{1}{}, ..., \vect{b}{k}{}]^T,  \vect{b}{j} \in  \mathbb{R}^d, \| \vect{b}{j}{}\| <= r.
\label{eq:basis}
\end{eqnarray}

It is important to mention that this set is arbitrary but fixed for all point clouds in the dataset. $r$ and $k$ are hyperparameters of the method, and $k$ can be used to determine the trade-off between computational complexity and the fidelity of the representation.


\paragraph{Feature calculation.} Next, we form a feature vector for every point cloud in a dataset by computing the minimal distance from every basis point to the nearest point in the point cloud under consideration:


\begin{eqnarray}
\mathbf{x}_i^{\mathbf{B}} = [\min_{\vect{x}{i}{j}\in X_i}{\dis{(\vect{b}{1},\vect{x}{i}{j})}}, \dots, \min_{\vect{x}{i}{j}\in X_i}{\dis{(\vect{b}{k}, \vect{x}{i}{j})}}]^T, \nonumber \\
\vect{x}{i}{}^{\mathbf{B}} \in \mathbb{R}^{k}. \label{eq:basis_dist}
\end{eqnarray}

Alternatively, it is possible to store the full directional information in the form of delta vectors from each basis point to the nearest point in the original point cloud:

\begin{eqnarray}
\mathbf{X}_i^{\mathbf{B}} = \big\{\big(\argmin_{\vect{x}{i}{j}\in X_i}{\dis{(\vect{b}{q}{},\vect{x}{i}{j})}} - \vect{b}{q}{} \big)\big\}  \in \mathbb{R}^{k \times d}, \label{eq:basis_points}
\end{eqnarray}

Other information about nearest points (\eg, RGB values, surface normals) can be saved as part of this fixed representation. The feature computation is illustrated in Fig.~\ref{fig:fig1}. The formulas (\ref{eq:basis_dist}) and (\ref{eq:basis_points}) give us fixed-length representations of the point clouds that can be readily used as input for learning algorithms.

\paragraph{BPS selection strategies.} We investigate a number of basis point selection strategies and provide details of these experiments in Sec.~\ref{subsection:selection}. Overall, random sampling from a uniform distribution in the unit ball provides a good trade-off between efficiency, universality of the generation process and surface reconstruction results, and we apply it throughout the experiments in this paper. Alternatively, an extensive 3D grid of basis points could be used in tandem with any existing 3D convolutional neural network in order to achieve maximum performance at the cost of increased computational complexity.


\paragraph{Complexity.} In this work, we use Euclidean distances between points for creating our encoding, but other metrics could be used in principle. Since we are working with 3D point clouds (which corresponds to having a small value for $d$), the nearest neighbor search can be made efficient by using data structures like ball trees \cite{omohundro1989five}. Asymptotically, $O(n\log{n})$ operations are needed for constructing a ball tree from the point cloud $X_i$ and $O(k\log{n})$ operations are needed to run nearest neighbor queries for $k$ basis points. This leads to an overall encoding complexity of $O(n\log{n} + k\log{n}$) per point cloud. The kNN search step can be also  efficiently implemented as part of an end-to-end deep learning pipeline \cite{kaiser2017learning}. Practically, we benchmark our encoding scheme for different values of $n$ and $k$ and show real-time encoding performance for values interesting for current real world applications. Please refer to the supplementary materials for further details.

\section{Analysis}
\label{section:analysis}
\subsection{Comparison to occupancy grids, TSDFs and plain point clouds}
\label{subsection:comparison}

\begin{figure}[t!]
\centering
    \includegraphics[trim={0.2cm 0cm 0.2cm 0cm},clip, width=\linewidth]{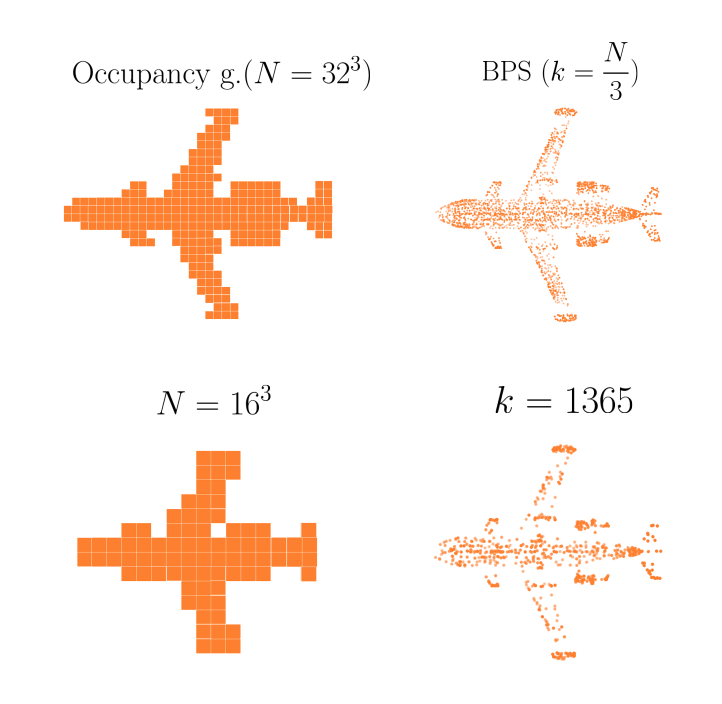}
\caption{%
\textit{Surface encoding with occupancy grids (left) and basis point sets (right).} With the same length of encoding $N$ our method can capture surface details more accurately. Even when using only $k \approx10^3$ basis points, our method can capture details of a surface (bottom right).}
\label{fig:comparison_qual}
\end{figure}

\paragraph{Informal intuition.} Compared to occupancy grids and TSDFs, the efficiency and superiority of the proposed BPS encoding is based on two key observations. First, it is beneficial for both surface reconstruction and learning to \emph{store some continuous global information} (\eg, Euclidean distance to the nearest point) in every cell of the grid instead of simple binary flags or local distances. In the latter case, most of the voxels remain empty and, moreover, the feature vector will change dramatically when slight translations or rotations are applied to an object. In comparison, every BPS cell always stores some information about the encoded object and the feature vector changes smoothly with respect to affine transformations. From this also stems the second important observation: when every cell stores some global information, we can use a much smaller number of them in order to represent the shape accurately, thus \emph{avoiding the cubical complexity of the extensive grid representation}. This can be seen in Fig.~\ref{fig:fig1} and bottom right Fig.~\ref{fig:comparison_qual}, where $k \approx n$ basis points are able to capture the outline of the original cloud. 

We will now validate this intuition by comparing the aforementioned representations in terms of surface reconstruction and actual learning capabilities.

\begin{figure}[t!]
\centering
    \includegraphics[trim={1.8cm 0cm 3.3cm 0cm},clip,width=\linewidth]{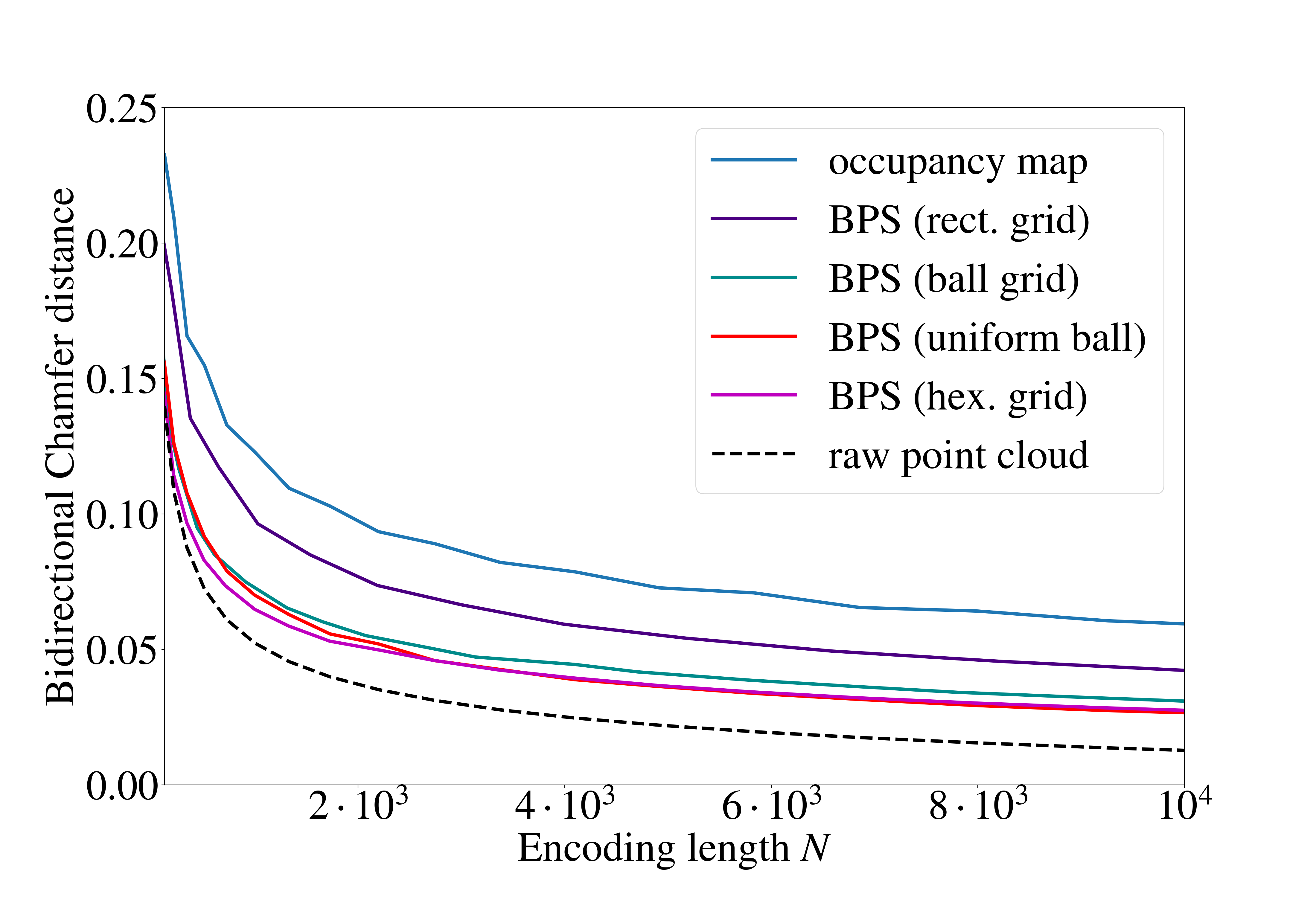}
\caption{%
\textit{Surface reconstruction quality vs. encoding length for different 3D data encoding methods.} We measured the Chamfer distance on $10^3$ encoded and reconstructed random shapes from the ModelNe40 dataset. The suggested representation is more accurate in representing surface details than standard occupancy grid. The performance of our best basis selection methods is close to encoding the surface with subsampled unordered point clouds while being a fixed-length representation that can be directly used with a wide range of machine learning algorithms. See Sec.~\ref{section:analysis} for further details.}
\label{fig:comparison}
\end{figure}

\paragraph{Surface reconstruction experiments.}  Independent of a certain point cloud at hand, how well does the encoding capture the details from the object? To answer this question, we take $10^3$ random CAD models from the ModelNet40~\cite{wu20153d} dataset and construct synthetic point clouds by sampling $10^4$ points from each surface. We compare three approaches of encoding the resulting point clouds: storing them as is (raw point cloud), occupancy grid and the proposed encoding via basis point sets as suggested in Eq.~\ref{eq:basis_points}.


For all methods we define a fixed allowed description length $N$ (as $N$ floating point values) and compare the normalized bidirectional Chamfer distance between the original point cloud $X$ and the reconstructed point cloud $X^r$ for the different encodings:

\begin{align}
\dis_{CD}(X, X^{r}) &= \frac{1}{|X|} \sum_{\vect{x}{i}{} \in X}{\min_{\vect{x^r}{i}{} \in X^r}{||\vect{x}{i}{}-\vect{x^r}{i}{}||^2}} \nonumber \\
&+ \frac{1}{|X^{r}|} \sum_{\vect{x^r}{i}{} \in X^r}{\min_{\vect{x}{i}{} \in X}{||\vect{x}{i}{}-\vect{x^r}{i}{}||^2}}.
\label{eq:chamfer}
\end{align}

With the same length of the description $N$ we can either store $N/3$ points from the original point cloud, $\sqrt[3]{N}\times\sqrt[3]{N}\times\sqrt[3]{N}$ binary occupancy flags or $N/3$ basis points with the matrix $\mathbf{X}_i^{\mathbf{B}}$ defined in Eq.~\ref{eq:basis_points}. From this matrix, a subset of original points can be reconstructed by simply adding corresponding basis point coordinates to every delta vector. For the occupancy grid encoding, we use the centers of occupied grid cells; please note that though a full floating point representation is not necessary to store the binary flag, in reality the majority of machine learning methods will work with floating point encoded occupancy grids and we assume this representation.

Fig.~\ref{fig:comparison} shows the encoding length and the reconstruction quality measured as Chamfer distance (\cf, Eq.~\ref{eq:chamfer}). The proposed encoding produces less than half of the encoding error compared to occupancy grids for point clouds up to roughly $10^4$ points (see Fig.~\ref{fig:comparison_qual} for a qualitative comparison). This is an indicator for its superiority for preserving shape information. The error curve for the basis point sets is close to the one of the subsampled point cloud representation.  The basis point set representation is less accurate than the raw point cloud since the resulting extracted points are not necessarily unique. However, the basis point set is an ordered, fixed-length vector encoding well-suited to apply machine learning methods.

\begin{figure}[t!]
\centering
    \includegraphics[trim={1.3cm 0cm 1.0cm 0cm},clip,width=\linewidth]{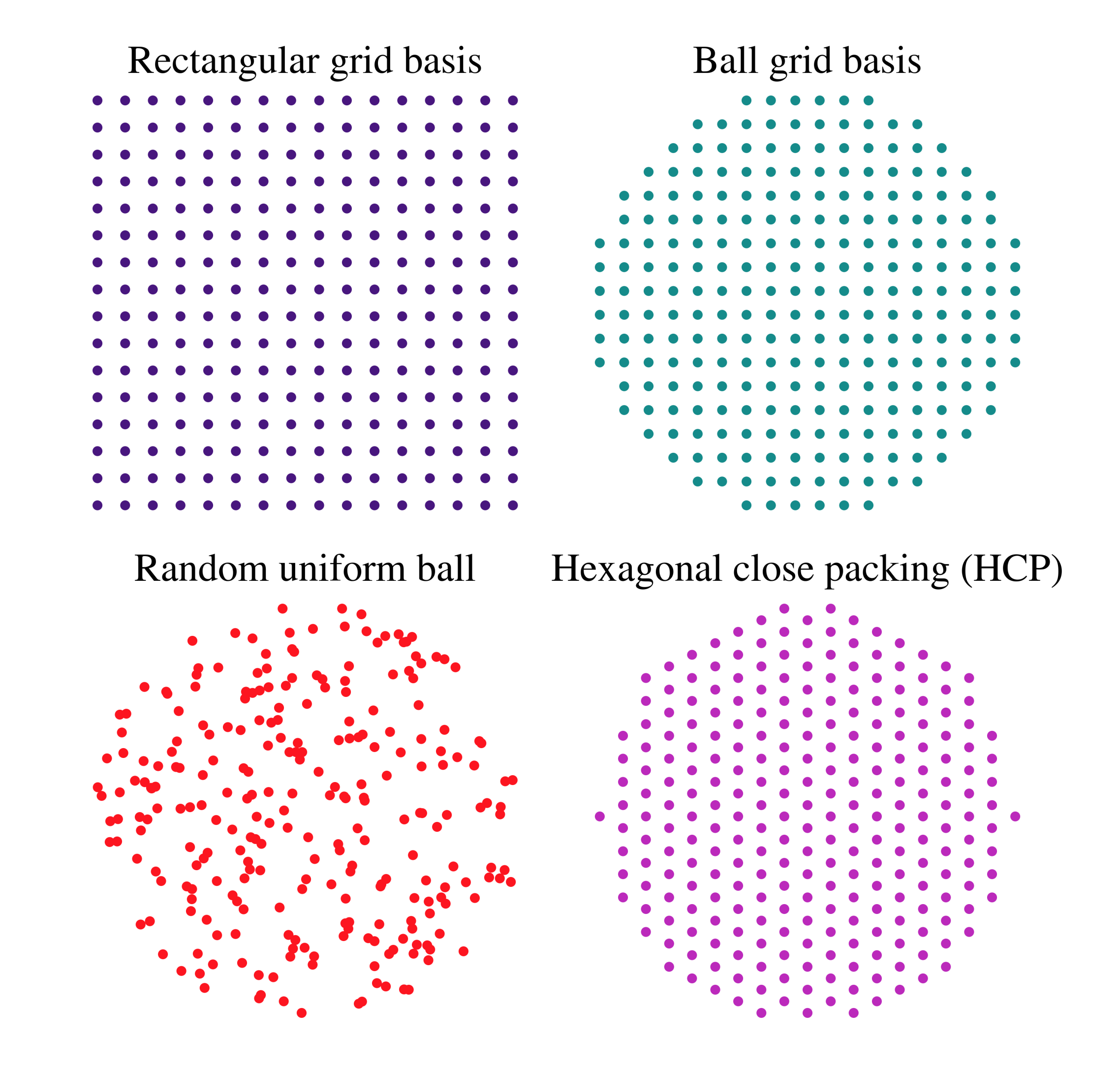}
\caption{%
\textit{Different basis point selection strategies.} See Sec.~\ref{subsection:selection} for details. In this work, we mainly use random uniform ball sampling for its simplicity and efficiency, as well as rectangular grid basis that allows us to apply 3D convolutions in a straightforward manner. Different BPS arrangements allow the usage of different types of convolutions.}
\label{fig:bps_selection}
\end{figure}

\subsection{Basis point selection strategies}
\label{subsection:selection}

We investigate four different variants of selecting basis points visualized in Fig.~\ref{fig:bps_selection}.

\paragraph{Rectangular grid basis.} A basic approach to basis set construction is to simply  arrange points on a rectangular $[-1, 1]^3$ grid. In that case, the basis point set representation resembles the truncated signed distance field  \cite{curless1996volumetric} representation. However, one important difference is that we do not truncate the distances for far-away basis points, allowing every point in the set to store some information about the object surface. We will show in Sec.~\ref{subsection:modelnet40} that this small conceptual difference has an important effect on performance. We are also allowing the full directional information to be stored in the cell as defined in Eq.~\ref{eq:basis_points}. Finally, BPS does not require the point clouds to be converted into watertight surfaces since \emph{unsigned} distances are used.

\paragraph{Ball grid basis.} Since all point clouds are normalized to fit in the unit ball by the transformation defined in Eq.~\ref{eq:unit_norm}, the basis points at the corners of the rectangular grid are located far away from the point cloud. These corner points in fact constitute $47.6\%$ of all the samples (this can be derived by comparing the volume ratio of a unit ball to a unit cube). Hence we can improve our sampling efficiency by simply trimming the corners of the grid and using more sampling locations within the unit ball.

\paragraph{Random uniform ball sampling.} One generic simple strategy to select points lying inside a $d-$dimensional ball is uniform sampling. This can be done by either rejection sampling from a $d-$dimensional cube or other efficient methods that are summarized in \cite{harman2010decompositional}.  
\paragraph{Hexagonal close packing (HCP).} We also experiment with \textit{hexagonal close packing} \cite{conway2013sphere} of  basis points. Informal intuition behind this point selection strategy is that it will optimally cover the unit ball with equally sized balls centered at the basis points \cite{hales2005proof}.

We show a comparison of reconstruction errors of $10^3$ ModelNet objects using the different sampling strategies in Fig.~\ref{fig:comparison}. Overall, the random uniform and HCP selection strategies provide the best reconstruction results. Using regular grids opens up possibilities for applying convolution operations and adds the possibility to learn translation and rotation invariant features.

We now evaluate the different encodings and basis point selection strategies with respect to their applicability with machine learning algorithms.



\section{Learning with Basis Point Sets}
\label{section:learning}
\subsection{3D Shape Classification}
\label{subsection:modelnet40}

\begin{center}
\begin{table}[t!]
  \centering
\resizebox{\columnwidth}{!}{
\begin{tabular}{llccc}
id & Method                        & acc.   & FLOPs              & params           \\
\hline
1 & VoxNet \cite{maturana2015voxnet} & $83.0\%$     & $>10^8$           & $9.0\times10^5$  \\
2 & Occ-MLP ($32^3$ grid) & $79.9\%\pm0.3$ & $3.4\times10^{7}$   &       $1.7\times10^{7}$      \\
3 & Occ-MLP ($8^3$ grid) & $74.5\%\pm0.2$  &   $1.1\times10^{6}$      &  $5.5\times10^{5}$  \\
4 & TDF-MLP ($32^3$ grid) & $80.0\%\pm0.3$  & $3.4\times10^{7}$   &       $1.7\times10^{7}$   \\
5 & TDF-MLP ($8^3$ grid) &  $75.9\%\pm0.3$  &  $1.1\times10^6$   &  $5.5\times10^5$   \\
6 & BPS-MLP ($32^3$ grid) &  $88.3\%\pm0.2$ &  $3.4\times10^{7}$       &   $1.7\times10^{7}$ \\
7 & BPS-MLP ($8^3$ grid) & $87.6\%\pm0.3$  &   $1.1\times10^{6}$      &  $5.5\times10^{5}$  \\
8 & BPS-MLP ($8^3$ ball)& $87.7\%\pm0.3$  &   $1.1\times10^{6}$      &  $5.5\times10^{5}$  \\
9 & BPS-MLP ($8^3$ rand)& $88.0\%\pm0.3$  &   $1.1\times10^{6}$      &  $5.5\times10^{5}$  \\
10 & BPS-MLP ($8^3$ HCP)& $88.1\%\pm0.3$  &   $1.1\times10^{6}$      &  $5.5\times10^{5}$  \\
11 & BPS-Conv3D ($32^3$ grid) & $89.8\%\pm0.2$ &   $3.5\times10^{8}$      &   $1.7\times10^{7}$ \\
12 & 9 $\rightarrow$ direct. vect.  & $86.2\%\pm0.3$   &   $2.2\times10^{6}$      &  $1.1\times10^{6}$  \\
13 & 11 $\rightarrow$ direct. vect.  & $90.8\%\pm0.3$   &       $3.8\times10^{8}$      &   $1.7\times10^{7}$ \\
14 & BPS-ERT \cite{geurts2006extremely} ($16^3$ g.) & $85.4\%\pm0.2$ &   N/A      &  N/A    \\
15 & BPS-XGBoost ($32^3$ g.) & $86.1\%\pm0.1$  &   N/A      &  N/A   \\
\end{tabular}}
\vspace{0.05cm}
  \caption{\textit{Comparison between occupancy grids, truncated distance fields (TDF) and BPS as input features for 3D shape classification on the ModelNet40~\cite{wu20153d} challenge.} We keep the model architecture fixed across experiments. Global BPS encoding significantly outperforms its local counterparts. See Sec.~\ref{subsection:modelnet40} for further details.}
  \label{table:comparison_learning}
\end{table}
\end{center}

One of the classic tasks to perform on point clouds is classification. We present results for this task on the \textit{ModelNet40}~\cite{wu20153d} dataset. We benchmark several deep learning architectures that use the proposed point cloud representation and compare them to existing methods that use alternative encodings. The dataset consists of ~$12\cdot 10^3$ CAD models from 40 different categories, of which $9.8\cdot 10^3$ are used for training. We use the same procedure for obtaining point clouds from CAD models as in~\cite{qi2017pointnet}, \ie, we sample $n=2048$ points from mesh faces, followed by the normalization process defined in Eq.~(\ref{eq:unit_norm}).

\paragraph{Comparison to occupancy grids and VoxNet.} To show the superiority of BPS features and to disambiguate contributions (\ie, the BPS encoding itself and the proposed network architectures), we fix a simple generic MLP architecture with 2 blocks of [fully-connected, relu, batchnorm, dropout] layers and perform training with $32^3$ rectangular grids of occupancy maps, truncated distance fields (TDFs) and BPS as inputs. 

Results are summarized in Tab.~\ref{table:comparison_learning}, rows 1-7. Using global distances as features instead of occupancy flags with the same network clearly improves accuracy, outperforming an architecture that was specifically designed for processing this type of input: VoxNet \cite{maturana2015voxnet} (row 1). TDFs store only local distances within the grid cell and suffer from the same locality problem as voxels (r.~4). It is also important to note that reducing the grid size affects these methods dramatically (rows 3 and 5, $5\%$ drop in accuracy), while the effect on the BPS is marginal (r.~6, $-0.7\%$). 

We also compare different BPS selection strategies in the rows 7-10 of Tab.~\ref{table:comparison_learning}.
In the absence of network operators exploiting the point ordering (e.g. 3D convolutions), random and HCP strategies give a slight boost in performance.
When the point order in a rectangular BPS grid is exploited with 3D convolutional deep learning models like VoxNet, performance improves at the cost of increased computational complexity (approximately two orders of magnitude more flops, Tab.~\ref{table:comparison_learning}, r.~11).

Substituting Euclidean distances with full directional information defined by Eq.~\ref{eq:basis_points} negatively affects the performance of a plain fully-connected network (Tab.~\ref{table:comparison_learning}, r.12) whereas it improves the performance of a 3D convolutional model (Tab.~\ref{table:comparison_learning}, r.~13).

To show the versatility of the proposed representation, we also use the same BPS features as input to an ensemble of extremely randomized trees (ERT~\cite{geurts2006extremely}) and XGBoost~\cite{chen2016xgboost} frameworks.


\paragraph{Comparison to other methods.}  Finally, we combine these findings with other enhancements (\eg, augmenting the data with few fixed rotations, improving learning schedule and regularization - please refer to the supplementary material and corresponding repository for further details) and compare our two best-performing models to other methods in Tab.~\ref{table:modelnet40}. 

In summary, simple fully connected network, trained on BPS features in several minutes on a single GPU, is reaching the performance of PointNet \cite{qi2017pointnet}, one of the most widely used networks for point cloud analysis. 3D-convolutional model trained on BPS rectangular grid is matching the performance of the PointNet++\cite{qi2017pointnet++}, while still being computationally more efficient. Finally, crude ensembling of 10 such models allows us to match state-of-the-art performance \cite{klokov2017escape} among methods working only on point clouds as inputs (\eg, without using surface normals that are available in CAD models but rarely in real-world scenarios).

\begin{center}
\begin{table}[t!]
  \centering
\resizebox{\columnwidth}{!}{
\begin{tabular}{lcccc}

Method                      & acc.     & FLOPs              & params           \\
\hline

RotationNet 20x \cite{kanezaki2018rotationnet}& $\bf{97.37}\%$ & \textgreater{$10^9$}  & $5.8\times10^7$ \\
MVCNN 80x  \cite{su2015multi}      & 90.1\%  & $6.2 \times 10^{10}$           & $9.9\times10^7$  \\
VoxNet \cite{maturana2015voxnet}           & 83.0\%   & \textgreater{$10^8$}            & $9.0\times10^5$  \\
Spherical CNNs \cite{esteves2018learning} & 88.9\%  & $2.9 \times 10^7$          & $5.0\times10^5$  \\
\hline
\multicolumn{1}{c}{\it{point cloud based methods:}}   \\
\hline
KD-networks \cite{klokov2017escape} & $\bf{91.8\%}$   &  \textgreater{$10^9$}         & \textgreater{$10^7$}  \\
KCNet \cite{shen2018mining} & 91.0\%   &  \textgreater{$10^8$}        & $9.0\times10^5$     \\
SO-Net \cite{li2018so} & 90.9\% &  \textgreater{$10^8$}           & \textgreater{$10^6$}  \\
DeepSets   \cite{zaheer2017deep}   & 90.0\%   &   $1.5\times 10^9$          &  $2.1\times10^5$         \\
PointNet++ \cite{qi2017pointnet++} &  90.7\%  & $1.6\times 10^9$            & $1.7\times10^6$  \\
PointNet   \cite{qi2017pointnet}     & 89.3\% & $4.4\times 10^8$            & $3.5\times10^6$             \\
PointNet(vanilla)   \cite{qi2017pointnet}  & 87.2\% &    $1.4\times 10^8$  & $8.0\times10^5$              \\
DeepSets (micro)  \cite{zaheer2017deep}    & 82.0\%  &  $3.8\times 10^7$           &      $2.1\times10^5$  \\
Ours (BPS-MLP) & 89.0\%  &  $\bf{7.6\times10^{5}}$    &    $3.8 \times10^{5}$ \\
Ours (BPS-Conv3D)  & $90.8\%$  &       $3.5\times10^{8}$      &   $4.4\times10^{6}$ \\
Ours (BPS-Conv3D, 10x)  & $91.6\%$  &       $3.5\times10^{9}$      &   $4.4\times10^{7}$ \\

\end{tabular}}
\vspace{0.25cm}
  \caption{\textit{Results on the ModelNet40~\cite{wu20153d} 3D shape classification challenge.} Simple fully connected network can be trained on BPS features in several minutes on a single GPU to reach the performance of PointNet.}
  \label{table:modelnet40}
\end{table}
\end{center}

\subsection{Single-Pass Mesh Registration from 3D Scans}

We showcase a second experiment with a different, generative task to demonstrate the versatility and performance of the encoding. For this, we pick the challenging problem of human point cloud registration. In this problem, correspondences are found between an observed, unstructured point cloud and a deformable body template.
\javier{we need to bring some related work from Rodola/Bronstein and others, pick from 3DCODED}
Traditionally, human point cloud registration has been approached with iterative methods~\cite{hirshberg2012coregistration, zuffi2015stitched}. However, they are typically computationally expensive and require the use of a  deformable  model at application time. Machine learning based methods~\cite{3dcoded} remove this dependency by replacing them with a sufficiently large training corpus. However, current solutions like~\cite{3dcoded} rely on multistage models with complex internal representations, which makes them slow to train and test. We encourage the reader interested in human mesh registration to review the excellent summary of previous work provided in~\cite{3dcoded}.

We use a simple DenseNet-like \cite{huang2017densely} architecture with two blocks (see Figure~\ref{fig:fig2}), where the input is a BPS encoding of a point cloud and the output is the \emph{location of each vertex} in the common template. Note that there is no deformable model in our system and that we do not estimate deformable model parameters or displacements; the networks learns to reproduce coherent bodies just based on its training data.
\label{subsection:faust}
\begin{center}
\begin{table}[t]
  \centering
\resizebox{\columnwidth}{!}{
\begin{tabular}{ccc}
Method                           & Intra (mms)    & Inter (mms)         \\
\hline
Stitched puppets   \cite{zuffi2015stitched} & 1.568  & 3.126\\
3D-CODED \cite{3dcoded} & 1.985  & 2.878\\
Ours  & 2.327 & 4.529\\
Deep functional maps \cite{FMNet} & 2.436 & 4.826\\
FARM \cite{FARM} & 2.81 & 4.123\\
Convex-Opt \cite{CONVEXOPT} & 4.86 & 8.304\\
\end{tabular}}
\vspace{0.25cm}
  \caption{\textit{Results for all published methods in the intra and inter challenge for the FAUST dataset, sorted by error in the intra challenge.} Our BPS-based network has a performance comparable to other methods while allowing single pass, real-time mesh registration, with no per-scan optimizations.}
  \label{table:faust}
\end{table}
\end{center}

To generate this training data, we use the SMPL body model~\cite{loper2015smpl}. SMPL is a reshapeable, reposable model that takes as input pose parameters related to posture, and shape parameters related to the intrinsic characteristics of the underlying body (\eg, height, weights, arm length).
We sample shape parameters from the CAESAR~\cite{Robinette2002} dataset, which contains a wide variety of ages, body constitution and ethnicities.
For sampling poses we use two sources: the CMU dataset~\cite{cmu} and a small set of poses inferred from a 3D scanner \christoph{Which set? Is it available? What is `small'? ;)}.
Since the CMU dataset is heavily populated with walking and running sequences,
we perform weighted sampling of poses with the inverse Mahalanobis distance from the sample to the CMU distribution as weight.\christoph{This could use more explanation}
We roughly align the CMU poses to be frontal. To increase the variation of the training data,
we introduce noise sampled from the covariance of all the considered poses to half of the data points.
From these meshes, a set of $10^4$ points is sampled uniformly from the surface of the posed and shaped SMPL template.
These point clouds are then used to compute the BPS encoding. We train the alignment network for 1000 epochs in only 4 hours and its inference time is less than 1ms on a non-GPU laptop.

To evaluate our method, we process the test set from the FAUST~\cite{bogo2014faust} dataset.
It is used to compare mesh correspondence algorithms by using a list of scan points in correspondence.
To find correspondences between two point clouds, we process each of them with our network,
obtaining as a result two registered mesh templates. The templates then define the dense correspondences between the point clouds.

\begin{figure*}[htb]
    \centering
    \includegraphics[width=\linewidth]{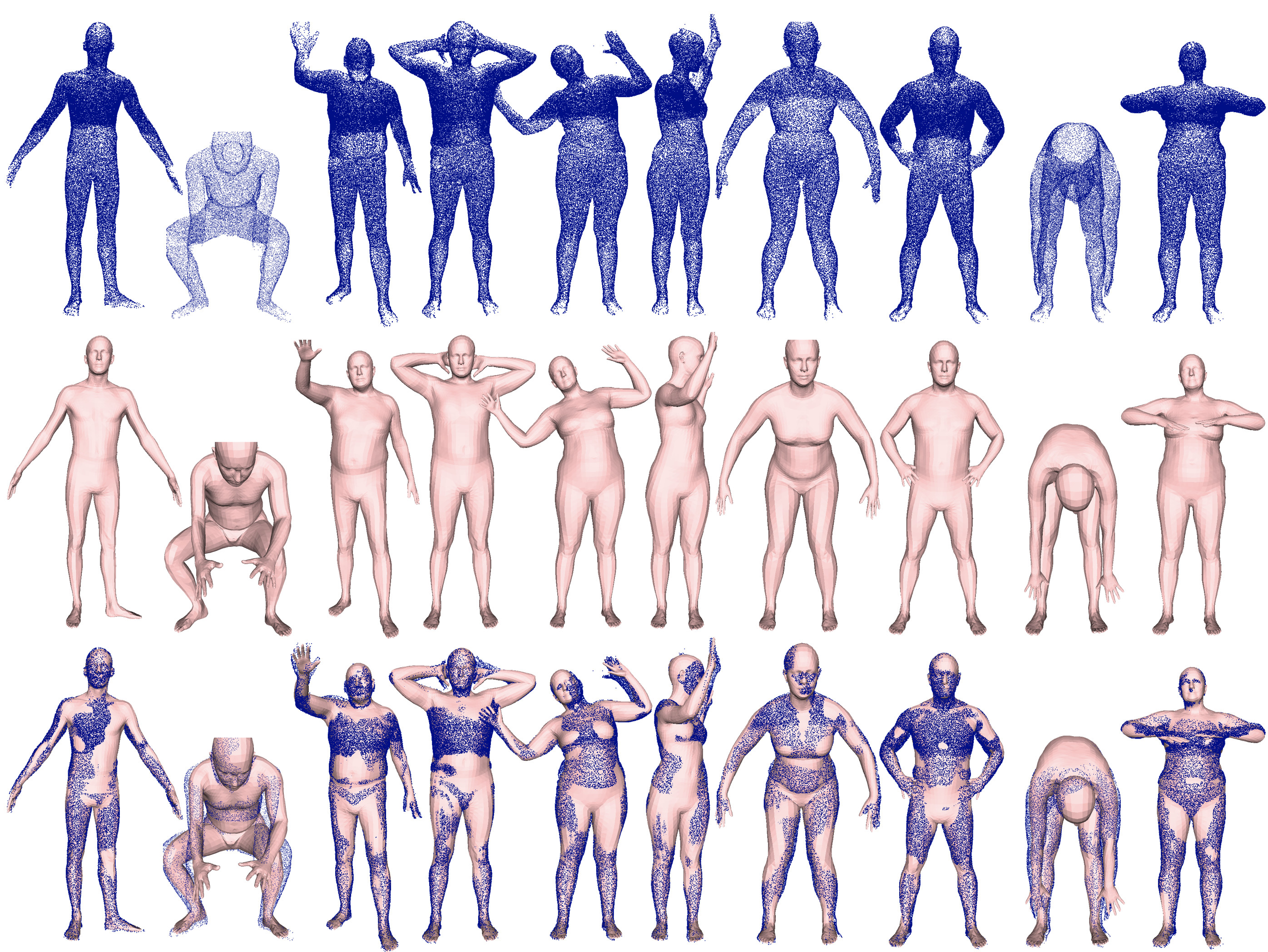}
    \vspace*{0.1cm}
    \caption{\textit{Point clouds of the FAUST dataset and the predicted meshes.} \textbf{Blue:} point cloud from a 3D scanner. \textbf{Skin color:} predicted mesh by our model through processing of its BPS representation. Note that the network produces the position of each output vertex; their coherent structure is learned solely from the training data.}
    \label{fig:faust}
\end{figure*}

We obtain an average performance of $2.327$mm in the intra-subject challenge and $4.529$mm in the inter-subject challenge (see Tab.~\ref{table:faust}).
These numbers are comparable, but higher than state-of-the-art methods like~\cite{3dcoded} or \cite{zuffi2015stitched}. However, we note
that the two methods outperforming BPS in the FAUST intra challenge are orders of magnitude slower than our system. The two-stage procedure in~\cite{3dcoded} takes multiple minutes and the particle optimization in~\cite{zuffi2015stitched} takes hours, while our system
produces alignments in ~1ms (for qualitative results, see Fig.~\ref{fig:faust}). This enables real-time processing of 3D scans, which was previously impossible, or can be used as a first step for faster multistage systems that refine the accuracy of this single stage method.
We also provide a qualitative evaluation on the Dynamic FAUST\cite{dfaust:CVPR:2017} dataset in the supplementary video \footnote{\url{https://youtu.be/kc9wRoI5JbY}}.


\section{Conclusion and Future Work}
In this paper, we introduced \textit{basis point sets} for obtaining a compact fixed-length represenation of point clouds. BPS computation can be used as a pre-processing step for a variety of machine learning models. In our experiments, we demonstrated in two applications and with different models the computational superiority of our approach with orders of magnitudes advantage in processing time compared to existing methods, remaining competitive accuracy-wise. 
We have shown the advantage of using rectangular BPS grid in combination with standard 3D-convolutional networks. However, in future work it would be interesting to consider other types of BPS arrangements and corresponding convolutions \cite{hoogeboom2018hexaconv, cohen2018spherical, esteves2018learning, graham2017submanifold} for improved efficiency and learning rotation-invariant representations.


%

\clearpage

{\small
\bibliographystyle{ieee_fullname}
\balance
\bibliography{egbib}
}



\section*{Supplementary material}

# Efficient Learning on Point Clouds with Basis Point Sets
## Supplementary Material

Sergey Prokudin *
Max Planck Institute for Intelligent Systems
Tübingen, Germany
sergey.prokudin@tuebingen.mpg.de

Christoph Lassner [†]
Amazon
Tübingen, Germany
classner@amazon.com

Javier Romero [†]
Amazon
Barcelona, Spain
javier@amazon.com

## 1. Encoding time

Encoding performance is crucial for real-world applications. Whereas we only state the algorithmic complexity $O(n \log n + k \log n)$ and note that encoding enables real-time application in the main paper, we provide a detailed analysis of its runtime in this section. For our benchmark, we take a random subset of $10^3$ ModelNet40 CAD models [10] and sample $10^1 - 10^5$ points from their surfaces. We also vary the number of the basis points $k$ used for the encoding. We investigate two implementations of the BPS encoding scheme: the first one uses the $k$ nearest neighbor implementation available in the sklearn scientific computing package [7]. Here, we use the ball-tree version of kNN search to achieve $O(n \log n + k \log n)$ complexity of the encoding. This version of encoding can be easily parallelized across multiple CPUs for different point clouds. As an alternative, we also provide a TensorFlow [1] implementation of a direct algorithm for computing pairwise distances. Despite its $O(kn)$ complexity, it shows remarkable performance even for large values of $k$ and $n$ due to the highly efficient execution on a GPU. Various hashing-based algorithms for kNN search designed specifically for 3D data could be used as well [3, 4].

Results are summarized in Fig. 1. Both implementations show super real-time encoding performance for the main application case considered in this paper (*i.e.*, point clouds containing $n < 10^5$ points encoded with $k < 10^4$ points). Combined with the proposed deep network, this allows real-time mesh registration from raw scans. Overall, the BPS encoding can be tailored to a specific platform and application via the efficiency-fidelity trade-off in varying $k$.

## 2. Space and time complexity

Apart from the FLOPs comparison on the ModelNet40 task, we also provide the inference times for a subset of point cloud processing networks with available implementations and compare them to our models. The results are

---

*work was done during internship at Amazon
[†]the last two authors were equally involved

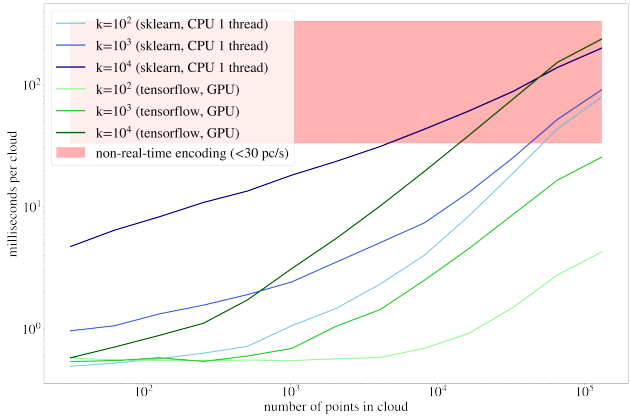

Figure 1. *BPS encoding time.* We measure encoding time per point cloud with respect to the number of input points, basis points $k$ and the implementation variant. Encoding can be done in real time for the main application cases considered in this paper (point clouds with $n < 10^5$, encoded with BPS of $k < 10^4$).

| Method | Inference time (CPU) | Inference time (GPU) | Model size |
|---|---|---|---|
| KD-networks [5] | 130ms | 41ms | 150MB |
| PointNet++ [9] | - | 22ms | 5.8MB |
| PointNet [8] | - | 8ms | 3.2MB |
| DeepSets [11] | 16ms | 0.5ms | **0.8MB** |
| Ours (BPS-MLP) | **0.04ms** | **0.04ms** | 1.6MB |
| Ours (BPS-Conv3D) | 56ms | 2ms | 18MB |

Table 1. *Inference time for different models.* Our fully connected model achieve sub-millisecond inference time on both, CPU and GPU.

presented in Tab. 1. Compared to other methods, our fully connected model is able to achieve sub-millisecond inference time even when being executed on a CPU.

## 3. Training details

A complete description of network architectures and training strategy will be made available through publication of our code repository. In the this section, we provide a description of the most important elements of the training.

**ModelNet40 models.** For the classification task, our MLP model consists of 2 fully connected hidden layers, each of size 400. Each layer is followed by a ReLU activation and dropout (with probability 0.8 and 0.4 repsectively). Class probabilities are given by a final dense layer with a softmax activation. The model is trained with the ADAM optimizer for 2500 epochs with a batch size of 2048 samples, with an initial learning rate of 1.0e-4 reduced to 1.0e-5 after 2000 epochs. Training data is augmented by few fixed-angle azimuth rotations.

Our Conv3D-model is similar to the original VoxNet [6] architecture, with 4 blocks of $3 \times 3 \times 3$ convolutions, with each pair of blocks followed by a max-pooling layer of size $2 \times 2 \times 2$. This is followed by 2 fully connected layers of size 512 (with separating dropout layers of 0.8 and 0.6). The model is trained with the ADAM optimizer for 505 epochs with a batch size of 512 samples, with an initial learning rate of 1.0e-4 reduced to 1.0e-5 after 500 epochs.

**FAUST model.** For the FAUST model, the architecture is depicted in Fig. 2 of the main manuscript. There, fully connected layers of size 1024 are used. The model is trained with SGD with momentum of 0.9 for 1000 epochs, with an initial learning rate of 1.0 and followup reductions by a factor of 0.5 every 5 epochs of no improvement on the validation set.

## 4. Results on DYNAMIC FAUST

The efficiency of BPS allows us to process large datasets efficiently. We showcase this in the supplementary video[1], where thousands of point clouds from the DYNAMIC FAUST dataset [2] are aligned with BPS. We use exactly the same network used in the FAUST experiments without any retraining or fine-tunning; this shows that our network generalizes well to unseen poses and subjects. Each frame is processed independently and no smoothness post-processing was applied to the video. The performance is pretty consistent across the sequences, with one negative factor impacting the accuracy of the alignments: the presence of strong outliers from the floor and the scanner. While the system is robust to the presence of few spurious points around the body, it is sensitive to large amounts of points far from it. Those large chunks drastically change the representation due to the size normalization of the point cloud, which makes the point cloud very different from any cloud in the training set. We believe this could be easily alleviated by introducing similar synthetic noise in our training data or making the normalization more robust to noise.

---

[1] https://youtu.be/kc9wRoI5JbY



\end{document}